# Underwater and Surface Aquatic Locomotion of Soft Biomimetic Robot Based on Bending Rolled Dielectric Elastomer Actuators

Chenyu Zhang, Chen Zhang, Juntian Qu, and Xiang Qian

*Abstract*— All-around, real-time navigation and sensing across the water environments by miniature soft robotics are promising, for their merits of small size, high agility and good compliance to the unstructured surroundings. In this paper, we propose and demonstrate a mantas-like soft aquatic robot which propels itself by flapping-fins using rolled dielectric elastomer actuators (DEAs) with bending motions. This robot exhibits fast-moving capabilities of swimming at 57mm/s or 1.25 body length per second (BL/s), skating on water surface at 64 mm/s (1.36 BL/s) and vertical ascending at 38mm/s (0.82 BL/s) at 1300 V, 17 Hz of the power supply. These results show the feasibility of adopting rolled DEAs for mesoscale aquatic robots with high motion performance in various water-related scenarios.

## I. Introduction

Versatile soft robots capable of locomotion in ambient aquatic environments are becoming desirable for long-term monitoring, all-round perception and exploration, from the top surface to the deep water [1], [2]. Inspired by natural animals, which evolved optimal body shapes along with strong motion abilities in various space, researchers introduced delicate structures and mechanisms to robots to realize multimodal locomotion and cross-interface operations [3], [4]. Moreover, simple, durable, all-purpose soft actuators and robots crossing bottom to surface environments are favorable for large-area water surveys [5].

Different soft actuation technologies have been applied to robotic swimmers. Fluidic actuators [6], [7] were widely explored and exhibited fast maneuver ability in water due to their large output force but were also limited by their pump system, bringing difficulty to on-board powering. Electrical driven soft actuators were another big family of artificial muscles, including ionic polymer-metal composites (IPMCs) [8], and DEAs [9], [10], [11]. Heat-driven biomimetic fish with shape memory alloys (SMAs) embedded were also presented in previous works [12], [13]. Since most energy was lost during thermal-mechanical conversion, there is few attempts of SMA robots for long-term usage. Of the existing

This work was supported by key technology program JSGG20201102065602006 of Shenzhen, China.

Chenyu Zhang, Chen Zhang, and X. Qian are with the Division of Advanced Manufacturing, Tsinghua Shenzhen International Graduate School, Tsinghua University, Shenzhen, China. {cy-zhang21, zchen22}@mails.tsinghua.edu.cn, qian.xiang@sz.tsinghua.edu.cn.

Chenyu Zhang, X. Qian are with the Open Faculty for Innovation, Education, Science, Technology and Art (Open FIESTA) Center, Tsinghua University, Shenzhen, China.

J. Qu is with the Institute for Ocean Engineering, Tsinghua Shenzhen International Graduate School, Tsinghua University, Shenzhen, China. (juntian.qu@sz.tsinghua.edu.cn)
.

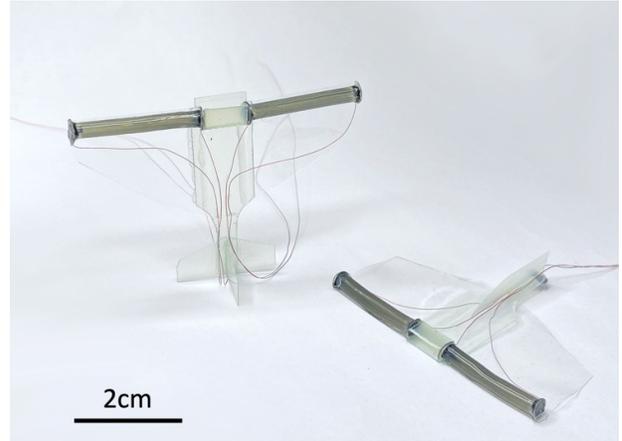

Figure 1. The soft aquatic robot with two rolled DEAs achieving flapping-fin movements at minimum fabrication.

propulsion methods, DEAs appeared to be promising in the field of centimeter scale robots in free water space for their large strain and fast response abilities [14].

he structural simplicity and minimal system requirements also extended DEAs' prospects. A fast-moving swimmer powered by DEA was developed [15], showing a maximum speed of 135 mm/s (1.45 BL/s) with fin-flapping frequency of 5 Hz under tethered locomotion. However, high voltage power supply over 6 kV was required for considerable pectoral deformation and thrust force, which limited the miniaturization of the whole system. In addition, pre-stretching and mold-casting techniques also added extra fabrication requirements.

To address these problems, one practical way is to stack multilayer dielectric elastomers instead of a whole piece of elastomeric sheet. With the thickness of each layer decrease to less than 35 $\mu$m, the required driving voltage could reduce to 1500V when driven by sinusoidal signals. Several DEAs-driven soft robots based on this principle with no need for pre-stretching were developed [10], [16], [17], opening the possibility of scalable and accessible fabrication process. To further advance the output force of the actuator, a compact rolled configuration of DEAs were proposed and well-modeled in recent years [18]. Despite their outstanding performance in the application of insect-scale flying robots [19], [20], few studies have been reported on rolled DEA-powered underwater robots. To our best knowledge, only one study of a fish-like robot was conducted [21], yet the maneuver capabilities were not fully exploited. In consideration of their similar characteristics to natural muscles, rolled DEA-powered soft vehicles that compete aquatic animals are expected.

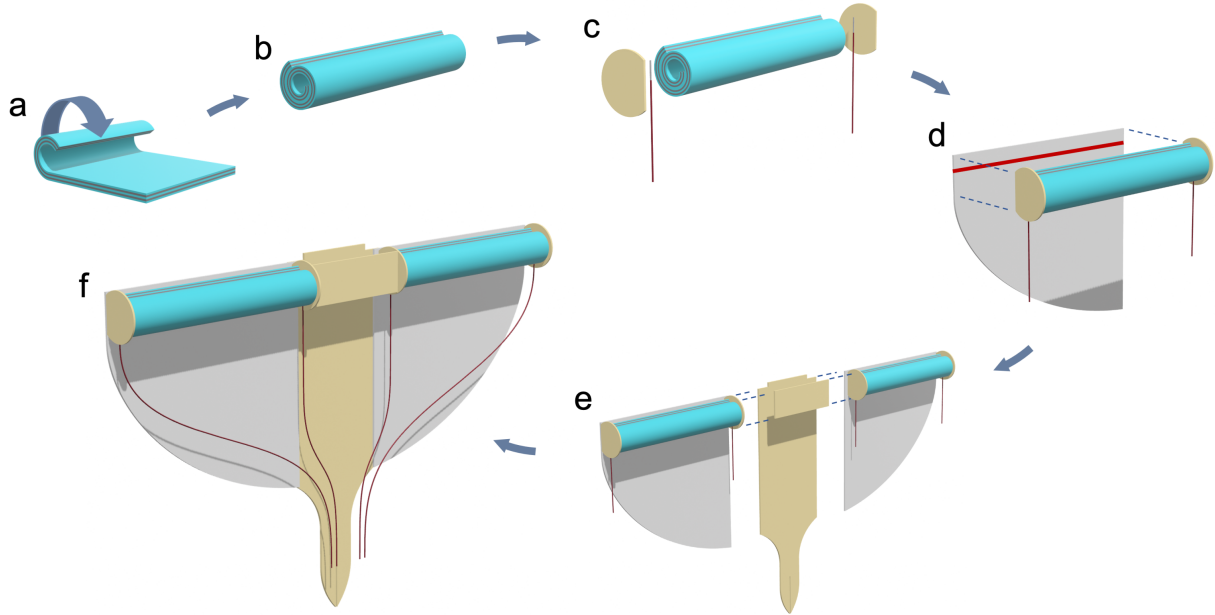

Figure 2. The fabrication process of the soft aquatic robot. (a) The DE multilayer was rolled into a cylinder (b). (c) Electrical wires and two insulation caps were fixed to the ends of tube by conductive carbon paint then formed a rolled DEA. (d) Each rolled DEA was bonded to a fin-shaped silicone film by silicone adhesive to form a bending actuator and flapping fin. The red line indicates the contact area of the tube and film where the glue was applied. (e) A pair of flapping fins were glued to a flexible chassis to finish the robot assembly (f).

Here we present a compact bending actuator based on rolled DEAs by adding a constraining membrane, resulting a considerable ~55 mN blocked force. The deflection styles of the actuator mimic the fin-flapping of marine animals like mantas, from which we built and demonstrate a simple-structured, but versatile soft robot for water-related purposes. The robot can perform underwater swim, top surface skating, and vertical lifting from the bottom at 1.25 BL/s, 1.36 BL/s, 0.82 BL/s, respectively. It is validated that the flapping-fin propulsion could be a common strategy of locomotion across the aquatic environment, from the bottom to surface.

## II. DESIGN AND FABRICATION

### A. Design Considerations

A general idea on designing a bending actuator is to adopt two shape-morphing units with ability to elongate or contract, composing an agonist-antagonist pair, just like the working mechanism of natural muscles. A series of bending DEAs that use planar DE multilayer as deformation modules have been developed in [9], [10], [16]. The planar actuator was adhered to a non-stretchable yet compliant membrane that guiding the multilayer to bend. However, the large length-to-thick ratio (>30) of DE multilayer endows with a very low bending modulus, which greatly limits the resulting output force, usually below 10 mN [22], [23].

On the other hand, bending actuators comprising two or three rolled DEAs, with the free ends of which were bonded together to a common endcap, were tested in a few works [17], [21]. This configuration does not considerably diminish the driving force, yet its deflection is not enough to generate a fin undulating movement with sufficient amplitude for appreciable thrust force. This is mainly because the DEA tubes were not adhered by their column sides, giving a large equivalent bending radius when actuated. Considering that the elongation of each DEA unit is only ~1mm, the resulting deformation angle is not sufficient in water scenarios.

Consequentially, we take into account both output force and deformation angle. In our design, a compact rolled DEA that provides adequate output force is bonded to a constraining film by its side. Film with much higher elastic modulus should be selected to effectively restrict the elongation along the bonded side of DEA tube. The extension of the rest part is not affected thus a bending motion is achieved. This approach takes advantages of the two designs mentioned above and requires only a single DEA for each bending unit which further simplified the structure and made it easier to control.

### B. Fabrication Process

The bending DEA was made up of a tubular-shaped DEA rolled from DE multilayer and a constraining layer (0.2mm PDMS film). Nine DE layers made from commercial silicone elastomer (Silikon Addition Farblos 5) composed the multilayer sheet (Fig.2 (a)). Each DE layer was fabricated from liquid precursor that was spin coated at 2000 rpm for 45 s to get a thickness of ~31 um, then sequentially underwent vacuum degassing, thermo-curing, CNT electrodes transferring, and post-transfer baking procedure in sequence which were well-instructed in previous work [20]. The parameters are listed in Table 1.

The DE multilayer with rectangular shape was cut out by a digital cutting machine (Silhouette CAMEO 4), and was manually rolled into a compact tube with ~4 mm diameter and 25 mm length, in Fig. 3(a). Enameled wires (0.10 mm diameter, Elektrisola Grade 3) were then connected to the two ends of the rolled tube using conductive carbon paint, as shown in Fig. 2(c) and Fig. 3(b). We used FR4 caps and applied instant adhesive to seal the end-cap joints in order to moderate the electrical shorting problem. The rolled DEA was bonded to a pre-cut elastomeric film (0.2mm PDMS) to finalize the fabrication of bending actuator, described in Fig. 2(d) and Fig. 3(c).

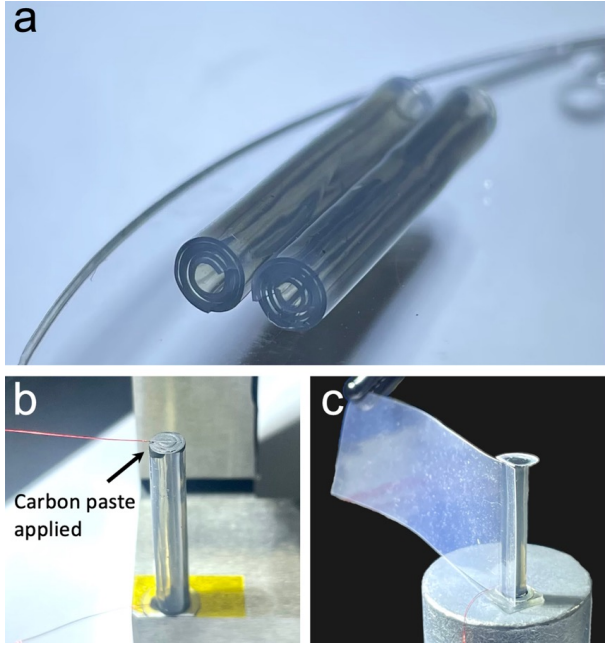

Figure 3. Key steps in the bending DEA fabrication. (a) DE multilayers were rolled into a tube with a diameter of ~4 mm followed by (b) wire connection and (c) bonding with soft constraining layer to finalize the fabrication of a compact bending DEA.

TABLE I. FABRICATION PARAMETERS OF DE MULTILAYER SHEET

| Layer No. | Vacuuming time (min) | Curing time at 70 ºC (min) | Post-transfer baking at 70 ºC (min) |
|---|---|---|---|
| 1 | 3.0 | 2.5 | 4.0 |
| 2 | 3.0 | 3.5 | 4.0 |
| 3 | 3.0 | 4.0 | 4.0 |
| 4 | 3.0 | 4.0 | 4.0 |
| 5 | 3.0 | 5.0 | 4.0 |
| 6 | 3.0 | 5.0 | 4.0 |
| 7 | 3.0 | 6.0 | 4.0 |
| 8 | 3.0 | 6.0 | 4.0 |
| 9 | 3.0 | 6.0 | 4.0 |

Finally, two bending DEAs that take the part of flapping fins were mounted to a flexible chassis (cut from 0.2 mm thickness FR4), shown in Fig. 2(e) and Fig. 2(f). Instant adhesive (Loctite 416) and silicone glue (Dow Corning 3140) were applied to ensure a reliable bonding. Enameled wires were carefully collected and attached to the rear of the chassis, with less constraint of the DEAs' flapping ends.

### III. CHARACTERIZATION OF BENDING DEAs

#### A. Experiment Setup

In the test of DEA bending angle, the fixed-end of the rolled actuator bonded to the chassis was gripped firmly to reduce the resonance effect of the robot body, though it contributes to the total amplitude of flapping motion either. We took the angular displacement at the free-end of the actuator when external voltage applied compared to its rest state, shown in Fig. 4. The powering wires were then connected to a high voltage supply (Agitek ATA-2082), which amplified the sinusoidal signal from a waveform generator (RIGOL) and served as the applied voltage.

For blocked force measurements, PTFE head was mounted to the force gauge probe (Mark-10 Series 7) and a tile-shaped block was used to constrain the flapping movement, which conformed to the tubular shape of the rolled DEA and helped to minimize the unwanted resistance and adhesion force.

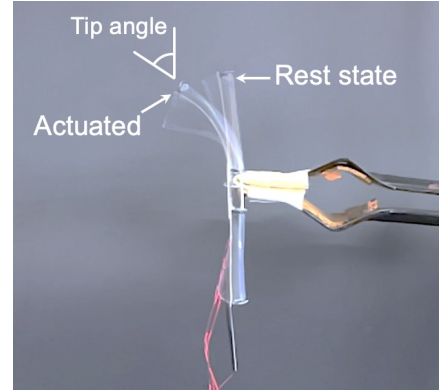

Figure 4. The actuated and rest state of the rolled DEA's bending modes. Deflection angle at the tip was used to describe the degree of curving. A maximum amplitude ~17mm at the free end was recorded.

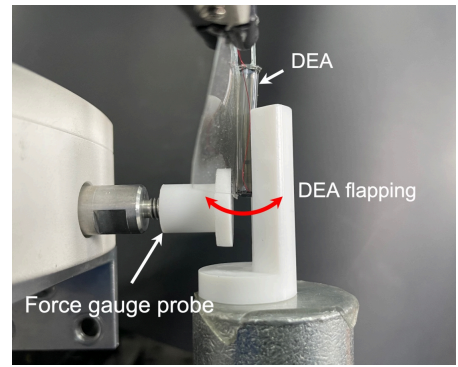

Figure 5. Experimental setup for measuring the blocked force in the flapping motion of DEA. Specially shaped PTFE holders were used to reduce the friction and vibration. The tested bending DEA was hanged upside down to eliminate the gravity effect.

The bending DEA was suspended and hold in order to eliminate the gravity contribution, as shown in Fig. 5.

## B. Results

Fig. 6 shows the bending angle as a function of frequency of with 1200-Vpp (peak-to-peak voltage) sinusoidal input. The flapping amplitude fluctuated around 46º as driving frequency varies from 3 Hz to 18 Hz, while the displacement at the tip drastically reduced over 20 Hz. This can be regarded as the response bandwidth of the current actuator. The maximum angle range of 52 ± 2º appeared at 17 Hz, representing a resonance point of bending modes; thus, in the following tests, driving frequency is simply set to 17 Hz to achieve a better performance. However, a more complicated mathematical model for the current design may be adopted in our future work to find a more appropriate resonance point with theoretical meanings.

Fig. 7 depicts the increment of flapping angle as the driving voltage climbs up. Note that an approximately linear relationship is displayed when power input is above 600V. This will contribute to the closed-loop control in the future. It is found that the actuator tends to buckle in the middle when undergoing an excessive bending stroke, induced by a voltage more than 1300 V (> 42 V/um of electric field applied to each elastomeric layer). We concerned the buckling effect would shrink the tube stiffness and not more promote the output force. So higher voltage is not tested, even though the dielectric limit of DE layer is still far to reach.

We measured the blocked force at 1200 V shown in Fig. 8. The bending DEA exhibits a ~55 mN blocked force and is sufficient to thrust a gram-scale robot in water. Though the rolled DEA's force produced under bending modes is merely a fraction of which generated by their linear modes, about 200 mN in [19] [20], it is still 6-8 times larger than planar-shaped bending DEA at the same scale [23]. It is supposed that the unevenness of the curve corresponds to the shaking from the DEA free-end and elastomeric fin.

These results show a vast perspective using rolled DEA as fast-response, high performance bending actuators. Given the merits of the actuator, soft robots with strong motion capabilities powered by bending DEAs are expected.

## IV. DEMONSTRATION OF AQUATIC LOCOMOTION

A series of experiments demonstrating the locomotion capabilities of the robot were conducted in a tank filled with purified water. All tests were open loop and operated under 1200 V, 17 Hz driving voltage. Fast move of underwater swimming, striding on surface, as well as vertical ascending were successfully performed and presented in Fig. 9, Fig. 10 and Fig. 11, respectively. Further demonstrations with details are showed in the supporting video. All the pictures and supporting videos were acquired by a smartphone at a resolution of 4K, with the DEAs' characterization videos captured at 240 fps and videos for locomotion demonstration captured at 60 fps. The speed of the locomotion was estimated by timing the robot as it moved over its entire body length (46 mm), then converting the result to absolute velocity.

In our current locomotion experiments, we used the same commercial available high voltage supply; however, it will be convenient to develop a on-board high voltage supply system in our future works. Thus, the current experiment results can validate our proposal of using compact bending DEAs achieving flapping motion for aquatic robots.

## A. Swimming

Although the fin-flapping amplitude was cut down by ~50% due to the drag of the fluid, our robot achieved a high swimming speed at 57 mm/s (1.25 BL/s), leveraged by the powerful bending DEAs (Fig. 9). We observed traveling waves along the undulating fins, which resembles the median-paired fin (MPF) mode of meter-scaled marine animals like stingrays and mantas, propelling itself by generating vortex behind the trailing edge of the fins. This indicates fin-fapping strategy can be well adapt to centimeter-sized soft robots.

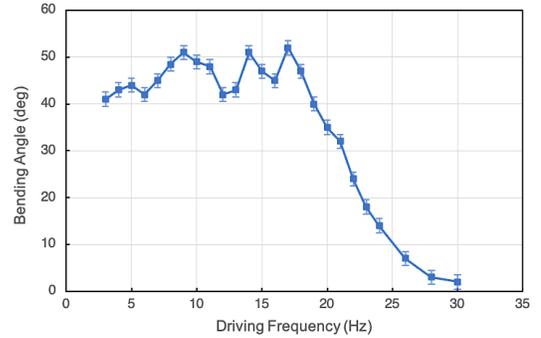

Figure 6. The bending angle of the tubular actuator under 1200 Vpp with respect to driving frequency from 3 Hz to 30 Hz with 1 Hz increment. Resonance occurs at ~9 Hz, ~14 Hz and ~17 Hz.

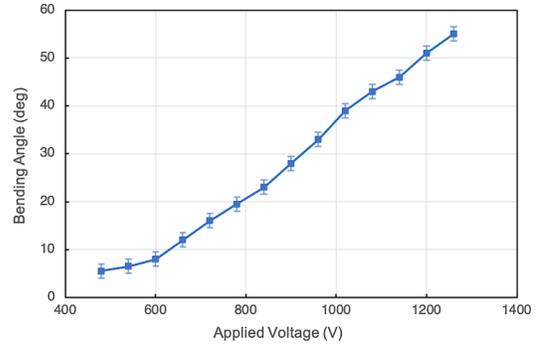

Figure 7. DEA's bending angle ascends as voltage increases from 480 Vpp to 1290 Vpp, at 17 Hz.

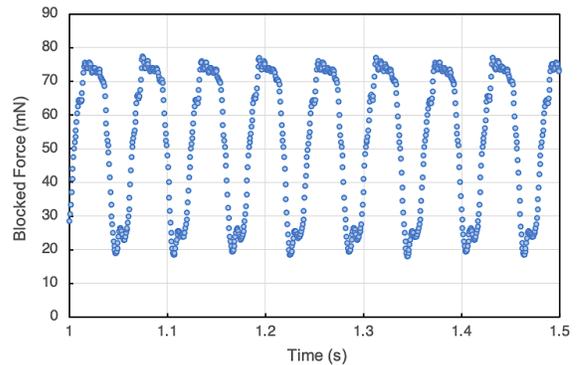

Figure 8. Sample blocked force of the bending motion of DEA driven at 17 Hz, 1200 Vpp.

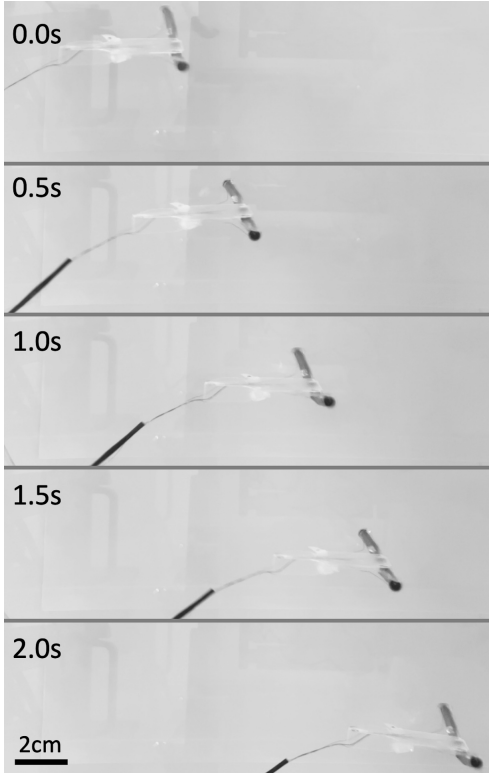

Figure 9. Sequence of underewater swimming of tethered robot (57 mm/s, 1.25 BL/s, driving at 1300 Vpp, 17 Hz). The sequence image also shows different phases of the flapping-fin in motion.

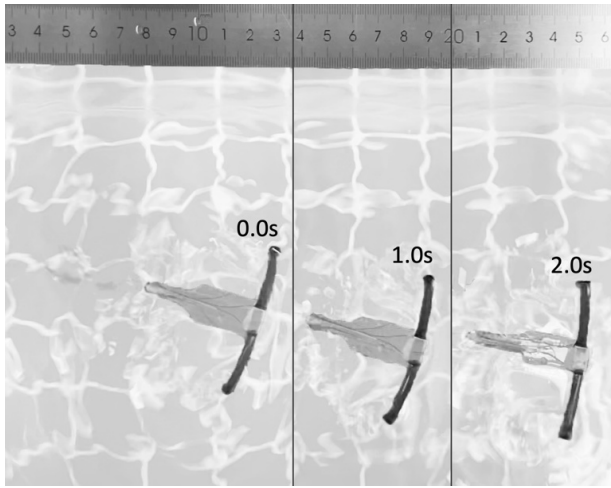

Figure 10. Sequence of the robot skating on water surface (64 mm/s, 1.36 BL/s). The fins kept unwetted during the motion due to their hydrophobic nature. A ruler was affixed closed to the water surface for direct measurement and comparison.

Note that the robot gently falls down as it swims forward, which means the structural imbalance and the powering wire slightly interfered the robot movements. Though it could be overcome by simply adding a float at an accurate place on the body, the lift force generated during fin-flapping is thought to be adequate to gravity. Hydrodynamic analysis should be developed to understand the swimming efficiency and minimize the drag of incoming flow in the future.

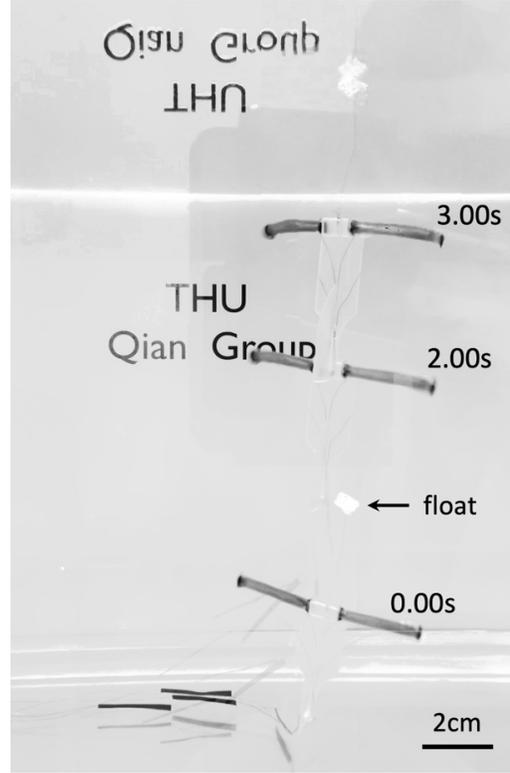

Figure 11. Sequence of underwater lifting of tethered robot (38 mm/s, 0.82 BL/s, driving at 1300 Vpp, 17 Hz).

### B. On-surface skating

Some insects, like water striders, mastering the surface tension and exhibit remarkable maneuvering abilities on water surface. Our soft robot accomplished on-surface skating with the help of the intrinsic hydrophobic property of the silicone fins, displayed in Fig. 10.

The top speed of the robot is slightly larger than of underwater swim by 12% (64 mm/s, 1.36 BL/s). This is thought to be the contribution of less water drag that generated at the leading edge (arc surface of the rolled DEAs) of the fins.

### C. Underwater taking lift

The robot realized a vertical ascending motion from the bottom to surface of water by connecting a tiny float (5 mm × 4 mm × 4 mm) to the chassis, without any adjustments on body structure. Ascending speed of 38 mm/s (0.82 BL/s) was observed (Fig. 11). Under the combined effect of buoyancy and gravity, the underwater flier was forced to stay vertically at the original state. Despite open loop, the robot passively maintained its upright position during the whole period of lifting up and dropping down.

## V. CONCLUSION AND FUTURE WORK

Soft robots propelled by flapping-fins possess outstanding moving capabilities. In this work, we proposed a novel design of quick-response bending actuator for flapping-fins from compact rolled DEAs, based on which we then developed a miniature soft swimmer. We demonstrated flapping-fin strategy is not limited to swim but also highly efficient on water surface and in vertical ascending, under identical structural design. The results show that the robot could be a promising candidate for real-time ecological tasks among the fastest soft aquatic vehicles.

In future work, a more comprehensive characterization of the bending actuator will be adopted and we will also further develop the robot attitude control as well as autonomous surface-water transition by new strategies, empowering the robot to carry out various applications across aquatic environment.

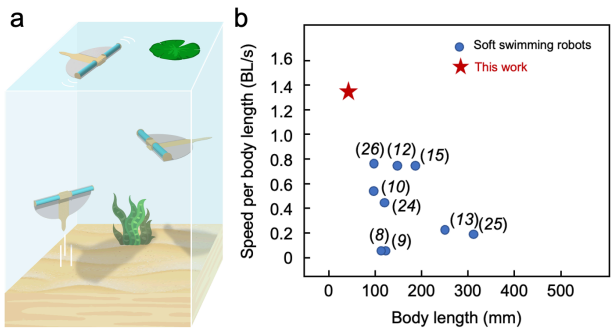

Figure 12. Conceptual image of the robot utilization and comparison of the robot motion capability with previous works. (a) Conceptual illustration shows the soft robot navigating aross the aquatic environment by swimming, vertical ascending, and skating on the surface. (b) Comparison of swimming speed (body length per second) versus body length between the proposed novel soft robot and the existing soft swimmers in literatures.

ACKNOWLEDGMENTS

The author would like to thank Y. Liu, D. Wang, Y. Chen, W. Liang, M. Zheng for their support and helpful discussions.